\documentclass[11pt]{article}
\usepackage{amssymb}
\usepackage{textcomp}
\usepackage{placeins}
\usepackage{array}

\usepackage{acl}

\usepackage{times}
\usepackage{latexsym}
\usepackage{multirow}

\usepackage[T1]{fontenc}
\usepackage[utf8]{inputenc}

\usepackage{microtype}

\usepackage{inconsolata}

\usepackage{graphicx}
\usepackage{booktabs}
\usepackage{url}

\usepackage{amsmath}

\title{CIVI: A Framework for Diagnosing\\Search Agent Failures in Civic Information}

\author{
  \textbf{Dingying Liu}\textsuperscript{1}\quad
  \textbf{Yunshun Zhong}\textsuperscript{2}\quad
  \textbf{Wentao Zhang}\textsuperscript{3}\quad
  \textbf{Yiyuan Li}\textsuperscript{4}
\\
  \textsuperscript{1}University of Sydney\quad
  \textsuperscript{2}University of Toronto\quad
  \textsuperscript{3}University of Waterloo\quad
  \textsuperscript{4}UNC-Chapel Hill
\\
  \texttt{dliu6951@uni.sydney.edu.au, yunshun.zhong@mail.utoronto.ca,}\\
  \texttt{w564zhang@uwaterloo.ca, yiyuanli@cs.unc.edu}
}

\begin{document}
\maketitle

\begin{abstract}
LLMs are increasingly deployed in public-sector settings, where incorrect guidance can cause irreversible harm. We introduce CIVI, the first framework for diagnosing search agent failures in civic information. Its benchmark instantiation jointly spans cross-national, inter-jurisdictional government contexts (federal, state, and local) and functional categories from an internationally adopted United Nations standard. We evaluate ten frontier search agents and find that none matches an attentive human baseline. Alongside accuracy, CIVI measures search invocation rate, selective no-search accuracy, and how often agents cite authoritative government sources. To perform this diagnosis, we introduce ARISE, which decomposes agentic search failures into four mutually exclusive modes, isolated via source-injection ablation. ARISE attributes 72.1\% of all observed failures to retrieval-bound causes rather than to gaps in the models' parametric knowledge.

\end{abstract}

\section{Introduction}

Frontier large language models (LLMs) are increasingly used in everyday decision-making \citep{pew2025chatgpt}, including in the public sector, where humans consult LLMs for guidance on benefit eligibility, tax obligations, and legal compliance. Unlike general factuality settings, authoritative public-sector information has near-zero tolerance for error. A single incorrect statement can produce significant and irreversible harm. New York City's MyCity chatbot was found to give illegal advice across regulatory domains. It told small business owners they have the right to fire employees who report harassment, advised restaurants that rat-bitten cheese was safe to serve, and offered similar misinformation on housing and labor matters \citep{lecher2024nyc, oecd2024mycity}. About 32\% of U.S. adults have turned to AI chatbots for health information \citep{kff2026aihealth}. At the same time, government agencies are deploying AI chatbots for public information at rapid scale: chatbots operated by the Internal Revenue Service, Social Security Administration, and Centers for Disease Control and Prevention have handled over 70 million human interactions, with several agencies seeing year-over-year growth of 5 times or more \citep{gao2026irsai, ssaoig2025telephone, castonguay2026cdc}. 

Evaluating LLM accuracy on authoritative public-sector information is therefore important, and what such an evaluation can establish depends on its scope and setting. Authoritative answers differ across countries and jurisdictional levels, so a model's overall civic reliability cannot be inferred from its accuracy in any single country. Humans obtain this information from general-purpose AI assistants that search the web as they answer, so a realistic evaluation must test models with agentic search rather than in a closed-book setup. While LLM factuality has been studied extensively in general settings \citep{min2023factscore, wei2024longform, manakul2023selfcheckgpt, li2023halueval, lin2022truthfulqa}, recent benchmarks targeting government information specifically remain confined to a single country at a single jurisdictional level \citep{majithia2026citizenquery, gao2025localbench}.  However, three gaps remain. First, no existing benchmark spans both cross-national and inter-jurisdictional government contexts with an international standard. Second, civic question answering (QA) evaluations report only aggregate accuracy, leaving diagnostic decomposition of failures unaddressed. Third, agentic search behavior, including whether models invoke search and cite authoritative sources, has not been systematically studied in civic QA. 

To address these gaps, we introduce CIVI (\textbf{C}ivic \textbf{I}ntergovernmental \textbf{V}erification of \textbf{I}nformation)\footnote{\raggedright Code: \href{https://github.com/dingyingliu/CIVI}{\texttt{github.com/dingyingliu/CIVI}}\\
Dataset: \href{https://huggingface.co/datasets/dingyingliu/CIVI}{\texttt{huggingface.co/datasets/dingyingliu/CIVI}}}, a framework for evaluating and diagnosing search agent failures in authoritative civic information across multiple governmental contexts. The present benchmark instantiation covers three English-majority federal democracies while the construction framework itself is country-agnostic.

Our main contributions are as follows:

\begin{enumerate}
\item \textbf{The CIVI Benchmark.} We instantiate CIVI as a benchmark of 4{,}097 expert-verified multiple-choice question-answer pairs grounded in 576 manually curated authoritative public-sector web pages. The dataset is stratified along three axes: four government-function categories drawn from the international Classification of the Functions of Government (COFOG) \citep{un2000cofog}, three jurisdictional levels (federal, state/provincial, municipal), and three federal democracies (Canada, the United States, and Australia). To our knowledge, it is the first benchmark to jointly span cross-national, inter-jurisdictional civic QA with an international classification standard.

\item \textbf{The ARISE failure diagnostic.} We introduce \textbf{A}gentic failu\textbf{R}e d\textbf{I}agno\textbf{S}tic prob\textbf{E} (ARISE), CIVI's failure-attribution procedure combining source-injection ablation with search-trace analysis to assign search agent failures to four mutually exclusive modes: search bypass, retrieval failure, grounding failure, and comprehension failure. Across all ten models, ARISE attributes 72.1\% of all observed failures to retrieval-bound causes rather than to gaps in parametric knowledge.
\item \textbf{Deployment-realistic evaluation.} We evaluate ten frontier LLMs across proprietary and open-source families, reporting overall accuracy alongside complementary measures of agentic search behavior, namely search invocation rate, Selective No-Search Accuracy, and a tier-inclusive Authoritative Hit Rate (AHR) measuring how often cited URLs come from authoritative government sources.
\end{enumerate}

\section{Dataset Construction}
\subsection{Data Source}
\label{sec:data-source}

\begin{table}[t]
\centering
\footnotesize
\setlength{\tabcolsep}{3pt}
\begin{tabular}{@{}l>{\raggedright\arraybackslash}p{3.8cm}r}
\toprule
Axis or unit & Levels or allocation & Count \\
\midrule
COFOG function & Public Order and Safety, Economic Affairs, Health, Social Protection & 4 \\
\addlinespace
Country & Canada, United States, Australia & 3 \\
\addlinespace
Jurisdictional level & Federal, state/provincial, local & 3 \\
\addlinespace
Stratification cells & \#COFOG $\times$ \#Country $\times$ \#Jurisdiction & 36 \\
\addlinespace
Source pages & 16 pages per cell & 576 \\
\addlinespace
Validated QA pairs & Approximately 114 QA pairs per cell & 4{,}097 \\
\bottomrule
\end{tabular}
\caption{CIVI stratification across COFOG function, country, and jurisdictional level.}
\label{tab:stratification}
\vspace{-15pt}
\end{table}
 
We construct CIVI from expert-curated authoritative public-sector web pages using a balanced three-axis stratification scheme: COFOG government function, country, and jurisdictional level. Table~\ref{tab:stratification} summarizes the resulting design.
 
\textbf{Government-function axis.} We adopt the Classification of the Functions of Government (COFOG), the international standard for classifying government expenditure by function, and sample from four categories: Public Order and Safety, Economic Affairs, Health, and Social Protection. These domains cover civic questions for which humans routinely seek authoritative information and where misinformation can produce material harm.
 
\textbf{Country axis.} We sample from three federal democracies: Canada, the United States, and Australia. This design tests whether models can answer civic questions across national institutional regimes rather than within a single-country setting.
 
\textbf{Jurisdictional-level axis.} Within each country, we sample source pages at three jurisdictional levels: federal, state or provincial, and local. This design captures the fact that authoritative civic answers often depend not only on the country but also on the level of government responsible for the policy, service, or legal requirement. The three axes yield 4$\times$3$\times$3 = 36 stratification cells.
 
\textbf{Source page curation.} Three domain experts with public-sector experience curated the source pages one cell at a time. For each cell, an expert first identified the government body or government-funded public-sector body responsible for that function and jurisdiction, and only then selected pages from its official or accredited website, reviewing many pages within each cell. For each cell this produced a candidate pool at least four times the 16-page target, drawn entirely from government and government-funded public-sector domains. From each pool, the experts retained the pages with the most substantive, decision-relevant content for residents, discarding landing pages and navigation hubs and excluding pages whose terms restrict use for AI training or evaluation. Trimming each pool to its 16 strongest pages yielded a final corpus of 576 source pages. Appendix~\ref{app:curation} details the full criteria, thresholds, and procedure.

\subsection{Question Generation}
\label{sec:question-design}
 
After curating the source pages described in 
\S\ref{sec:data-source}, we generate CIVI
QA pairs using Claude Opus 4.6 
\citep{claudeopus46}. We adopt a multiple-choice 
format, following factual-knowledge 
benchmarks \citep{hendrycks2021, clark2018, 
lin2022truthfulqa} and domain-specific factuality 
benchmarks \citep{pandit2025, jansen2025}. This format enables automated evaluation at scale. Each question must satisfy four criteria.
 
\textbf{Source derivability.} Each question must be 
directly answerable from a single source page, 
without external knowledge or cross-page synthesis. 
This single-page grounding also supports the 
source-injection diagnostic in \S\ref{sec:cafda}.
This criterion reflects the task-oriented structure promoted by the Canada.ca Content and Information Architecture Specification, the Australian Government Style Manual and Digital Service Standard, and U.S. federal web standards under the 21st Century IDEA: resident-facing service pages are generally designed to consolidate eligibility, procedures, fees, and deadlines for a specific task. 
 
\textbf{Concrete consequence.} Each question tests 
information whose accuracy carries a concrete 
consequence for a human: either a required action 
(eligibility, deadlines, fees, application steps, 
contact channels) or a substantive feature of a 
public service (how it operates, what it delivers, 
how its performance is measured). Generated questions 
are distributed across both types. Trivia, definitions, and common-knowledge questions 
are excluded.
 
\textbf{Jurisdictional precision.} Each question concerns a single specific jurisdiction and has exactly one correct answer for that jurisdiction.

\textbf{Source attribution.} Alongside each correct 
answer, the generator returns the verbatim 
quotations from the source page that license it. 
This constrains generation toward extractive 
answering and gives the \S\ref{subsec:validation} validators the precise span to verify. 
 
Each question presents eight options, of which one 
or more are correct. The remaining options are 
distractors. Distractors come from two sources: 
controlled perturbation of page content (changing a 
threshold, flipping a condition, altering a deadline, 
substituting an entity, or reordering steps) and 
plausible misconceptions a reader might draw from 
related civic knowledge but that the source page 
does not support. Distractors are deduplicated so 
that no two test the same underlying misconception. 
The full prompt is provided in 
Appendix~\ref{app:gen-prompt}.

\textbf{QA pair contents.} Each finalized QA pair 
records the question with its eight options, the 
correct option set, the gold source URL and page 
title of the authoritative source page, and the 
verbatim supporting quotations that license the 
answer. Each pair is also tagged with its COFOG 
category and jurisdictional level. The gold source 
URL serves as the provenance anchor for the 
\S\ref{sec:cafda} diagnostic, which checks whether 
a model retrieved the gold source page. The generated candidates are then passed to the validation procedure described in \S\ref{subsec:validation}.

\subsection{Human Validation}
\label{subsec:validation}

Two researchers with domain expertise in public-sector AI independently validated all $4{,}354$ generated pairs. Each reviewed the full dataset, so every pair received two independent annotations. The validators brought complementary experience in public-sector AI. One has multiple years of experience working with public-sector agencies on deployed retrieval-augmented systems, and the other has advised municipal government stakeholders on AI accuracy in civic information access.

For each pair, the annotator examined the question, re-examined the gold source page it was generated from, confirmed that the supporting quotations appear in that page, and verified the pair against the four \S\ref{sec:question-design} criteria: source derivability, concrete consequence, jurisdictional precision, and source attribution. Pairs flagged by either annotator as failing any criterion were excluded from the final dataset. Both annotators returned matching verdicts on 96.8\% of pairs (raw agreement), and 94.1\% of generated pairs were accepted by both annotators and retained in the final dataset. This acceptance rate confirms the effectiveness of our automated question generation pipeline.

\section{Evaluation Protocol}
\label{sec:protocol}

\subsection{Evaluation Setup}
\label{subsec:eval-setup}

Our agentic condition reflects how people actually access civic information. They query a publicly available AI assistant that retrieves authoritative sources at inference time, rather than being supplied curated context in advance~\citep{jo2023deploying}.

To standardize this setting across models, we route all agentic search through Exa, which provides a common search substrate and enables direct comparison of search behavior. Following the retrieval budgets used in prior agentic search evaluation~\citep{li2025searcho1, jin2025searchr1, zheng2025deepresearcher, song2025r1searcher}, each model may issue up to three rounds of search queries, with each round allowing unlimited search queries, and may retrieve a maximum of 10 URLs across queries. This range falls within the per-turn retrieval patterns of consumer AI assistants such as Claude~\citep{anthropic2025websearch} and ChatGPT~\citep{openai2024search}, both of which rewrite user prompts into multiple bounded search invocations. The full setup of our implementation is shown in Appendix~\ref{app:implementation}.

\subsection{Models}
\label{subsec:models}

We evaluate ten frontier LLMs selected to reflect the deployments most likely encountered by humans accessing public-sector information. Seven are closed-source, the consumer-facing default of each major provider (Gemini 3 Flash~\citep{gemini3flash}, GPT-5.2 Chat~\citep{gpt52chat}, and Claude Sonnet 4.6~\citep{claudesonnet46}) together with higher- and lower-tier variants from the same three families (GPT-5.4~\citep{gpt54}, GPT-5.4-mini~\citep{gpt54mini}, Gemini 3.1 Pro~\citep{gemini31pro}, and Claude Haiku 4.5~\citep{claudehaiku45}). The remaining three are open-weight models from major open-source families, including DeepSeek-V3.2~\citep{deepseekv32}, Mistral Large 3~\citep{mistrallarge3}, and Qwen3.6-Plus~\citep{qwen36plus}.

\subsection{Metrics}
\label{subsec:metrics}

We evaluate model responses with four metrics. Accuracy captures whether a model answers correctly. The other three, search invocation rate, Selective No-Search Accuracy, and authoritative hit rate, characterize the model's search and sourcing behavior, which the agentic deployment setting makes consequential beyond correctness alone.

\textbf{Accuracy.} We compute accuracy via exact set match (order-invariant and case insensitive), treating a response as correct only when it selects exactly the correct option set, with no missing or incorrect selections. Exact match aligns with civic information stakes, where partially correct answers can still produce materially misleading guidance.

\textbf{Search invocation rate.} The proportion of all
evaluation questions on which a model invokes search at
least once, reported both overall and separately by jurisdictional level.

\textbf{Selective No-Search Accuracy.}
The proportion of correct answers among questions for which the model did not invoke search. Because this subset is selected by the model and varies in size and composition across models, Selective No-Search Accuracy is a per-model behavioral diagnostic rather than a direct cross-model measure of closed-book ability. For cross-model comparison of search-policy calibration, we define $\Delta_{\mathrm{skip}}$ as Selective No-Search Accuracy minus accuracy when search is disabled on the full benchmark. Positive values indicate that a model preferentially skips search on questions that are easier for it, values near zero indicate little selection advantage, and negative values indicate that it skips search on comparatively difficult questions.

\textbf{Authoritative hit rate.} Models cite sources of varying authority, from official government domains to community forums, news outlets, and editorial sites, but only government domains can definitively answer institutional questions about benefits, eligibility, and procedural requirements. We define the authoritative set for a question based on its jurisdictional level: a federal question admits only federal government domains, a state question admits state and federal domains, and a municipal question admits municipal, parent-state, and federal domains.\footnote{For example, a question about San Francisco admits domains from the city of San Francisco, the State of California, and the federal government of the United States.} For each question, we compute the proportion of the model's cited URLs that come from authoritative government domains. The mean Authoritative Hit Rate (meanAHR) is the average of these per-question proportions across all questions for which the model cited at least one URL. Let $Q_m$ be the set of questions for which model $m$ cited at least one URL, let $n(q,m)$ be the number of unique URLs cited by $m$ in response to question $q$, and let $n_{\text{auth}}(q,m)$ be the number of those URLs from authoritative domains. The mean AHR for model $m$ is: 
\begin{equation}
\text{meanAHR}_m = \frac{1}{|Q_m|} \sum_{q \in Q_m} \frac{n_{\text{auth}}(q,m)}{n(q,m)}.
\end{equation}
We report 95\% confidence intervals (CIs) on per-model meanAHR using non-parametric bootstrap resampling with 1000 resamples over the per-question AHR values.

\section{CIVI}
\label{sec:civi-eval}

\subsection{Results}
\label{subsec:results}

We evaluate ten frontier LLMs on the full CIVI set of $4{,}097$
questions under the agentic search condition described in \S\ref{subsec:eval-setup}.
Table~\ref{tab:civi-eval-accuracy} reports aggregate exact-match
accuracy. As a comparator, we establish a \textbf{human
baseline} from $5$ participants, each answering the same 150
stratified questions with unrestricted search-engine access and unlimited time. Mean accuracy across these users is \textbf{92.67\%}, with a standard deviation of 3.1 percentage points. Model scores and rankings are largely stable across three runs on a stratified subset of 410 questions (Appendix~\ref{app:stability}).

\paragraph{All models trail the attentive human baseline.}
The strongest model, Gemini 3.1 Pro at 78.28\% (95\% CI \textpm 1.3 pp),
falls 14.4 percentage points below the human baseline. Despite access
to agentic search, no frontier LLM matches an attentive human on
civic question answering. On the same subset answered by the human participants, Gemini 3.1 Pro scores 77.33\%, 15.3 percentage points below the human baseline (Appendix~\ref{app:matched-human}).

\begin{figure}[tb]
\vspace{-10pt}
  \centering
  \includegraphics[width=\linewidth]{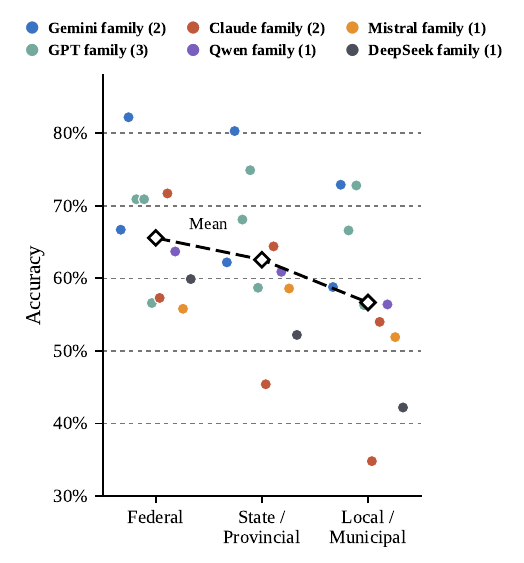}
  \caption{Accuracy declines as jurisdictional level 
descends, with mean accuracy falling from 65.6\% at 
federal questions to 56.7\% at local/municipal 
questions. Each dot is one model's accuracy at one 
level ($n=1{,}310$ federal, $n=1{,}334$ 
state/provincial, $n=1{,}453$ local/municipal). The 
dashed line traces the cross-model mean, and dots are 
colored by model family.}
  \label{fig:strip-plot}
  \vspace{-5pt}
\end{figure}

\paragraph{Performance declines with jurisdictional level.}
As Figure~\ref{fig:strip-plot} shows, mean accuracy declines monotonically from 65.6\% at the federal level to 62.6\% at the state/provincial level and 56.7\% at the local/municipal level, a difference of 8.9 percentage points between the federal and local levels. We hypothesize that this pattern aligns with the unequal representation of authoritative sources across general-purpose training corpora, with local sources especially sparse.

\begin{table}[t]
  \centering
  \begin{tabular}{lr}
    \toprule
    Model & Accuracy \\
    \midrule
    Human Baseline  & \textbf{92.67} \\
    \midrule
    Gemini 3.1 Pro    & \textbf{78.28} (\textpm 1.3) \\
    GPT-5.4           & 72.88 (\textpm 1.4) \\
    GPT-5.2 Chat      & 68.46 (\textpm 1.4) \\
    Claude Sonnet 4.6 & 63.05 (\textpm 1.5) \\
    Gemini 3 Flash    & 62.44 (\textpm 1.5) \\
    Qwen 3.6 Plus     & 60.20 (\textpm 1.5) \\
    GPT-5.4-mini      & 57.18 (\textpm 1.5) \\
    Mistral Large 3   & 55.33 (\textpm 1.5) \\
    DeepSeek V3.2     & 51.12 (\textpm 1.5) \\
    Claude Haiku 4.5  & 45.45 (\textpm 1.5) \\
    \bottomrule
  \end{tabular}
  \caption{Aggregate accuracy on CIVI. The strongest model's full benchmark score is 14.4 percentage points below the human subset baseline. On the matched subset, the gap is 15.3 percentage points (Appendix~\ref{app:matched-human}). Parenthetical values are 95\% Wilson score confidence interval half-widths in percentage points. The human baseline is the mean accuracy of five users with unrestricted search-engine access and unlimited time.}
  \label{tab:civi-eval-accuracy}
  \vspace{-3pt}
\end{table}

\paragraph{Inter-model spread widens at lower levels.}
The range of accuracies across models grows as jurisdictional level
descends. Federal accuracies span 26.4 percentage points (55.8\% to 82.2\%), state
accuracies span 34.9 percentage points, and municipal accuracies span 38.1 percentage points. Models
that perform similarly on federal questions diverge more sharply on
questions whose authoritative sources are sparser.

\paragraph{Cross-jurisdictional robustness varies dramatically.}
The federal-to-municipal accuracy drop varies widely
across models, from $+1.9$ percentage points for GPT-5.4, which improves
at the municipal level, to $-22.5$ for Claude Haiku 4.5. The GPT family is
flat across levels, with all three dropping less than
5 percentage points. We hypothesize that their substantially increased search invocation at lower jurisdictional levels contributes to this robustness in accuracy (\S\ref{subsec: agentic search}).

\paragraph{Top model lead widens at higher jurisdictional levels.}
The accuracy gap between the top-ranked model and the second-best widens substantially as jurisdictional level rises. At state, Gemini 3.1 Pro leads by 5.4 percentage points (80.3\% versus GPT-5.4 at 74.9\%). At federal, the lead widens to 10.5 percentage points (82.2\% versus Claude Sonnet 4.6 at 71.7\%), nearly double the state-level gap.

\paragraph{Accuracy varies less across countries than across jurisdictional levels.}
Averaging the ten model scores equally, accuracy is 63.4\% on U.S.
questions, 62.0\% on Australian questions, and 59.1\% on Canadian
questions. The range across countries is 4.3 percentage points,
compared with an 8.9 percentage point difference between the federal
and local levels (Appendix~\ref{app:country-results}).

\FloatBarrier

\begin{table*}[t]
\centering
\scriptsize
\setlength{\tabcolsep}{2.8pt}
\resizebox{\textwidth}{!}{%
\begin{tabular}{lccc@{\hspace{5pt}}cccccc}
\toprule
\multirow{2}{*}{\textbf{Model}}
& \multicolumn{3}{c}{\textbf{Accuracy}}
& \multicolumn{4}{c}{\textbf{Search rate}}
& \multirow{2}{*}{\shortstack{\textbf{Selective No-Search}\\\textbf{Accuracy}}}
& \multirow{2}{*}{\shortstack{\textbf{$\Delta_{\mathrm{skip}}$}\\\textbf{(pp)}}} \\
\cmidrule(lr){2-4}\cmidrule(lr){5-8}
& \textbf{Agentic}
& \textbf{Disabled}
& \textbf{RAG}
& \textbf{Overall}
& \textbf{Federal}
& \textbf{State/Prov.}
& \textbf{Local/Mun.}
& {}
& {} \\
\midrule
GPT-5.4
& \textbf{72.88} (\textpm 1.4)
& 60.4 (\textpm 1.5)
& 61.9 (\textpm 1.5)
& 69.8\% & 40.4\% & 73.0\% & 93.3\%
& 68.4\% & $+8.0$ \\

GPT-5.4-mini
& \textbf{57.18} (\textpm 1.5)
& 45.1 (\textpm 1.5)
& 47.4 (\textpm 1.5)
& 69.1\% & 41.7\% & 71.6\% & 91.6\%
& 53.3\% & $+8.2$ \\

GPT-5.2 Chat
& \textbf{68.46} (\textpm 1.4)
& 55.1 (\textpm 1.5)
& 52.4 (\textpm 1.5)
& 47.4\% & 19.1\% & 38.5\% & 81.1\%
& 66.4\% & $+11.3$ \\
\midrule

Claude Sonnet 4.6
& \textbf{63.05} (\textpm 1.5)
& 56.9 (\textpm 1.5)
& 57.7 (\textpm 1.5)
& $\geq$99\% & $\geq$99\% & $\geq$99\% & $\geq$99\%
& N/A & N/A \\

Claude Haiku 4.5
& \textbf{45.45} (\textpm 1.5)
& 33.3 (\textpm 1.4)
& 32.9 (\textpm 1.4)
& $\geq$99\% & $\geq$99\% & $\geq$99\% & $\geq$99\%
& N/A & N/A \\
\midrule

Gemini 3.1 Pro
& \textbf{78.28} (\textpm 1.3)
& 73.2 (\textpm 1.4)
& 68.4 (\textpm 1.4)
& 62.4\% & 41.1\% & 58.0\% & 85.5\%
& 78.0\% & $+4.8$ \\

Gemini 3 Flash
& 62.44 (\textpm 1.5)
& \textbf{66.5} (\textpm 1.4)
& 54.2 (\textpm 1.5)
& 13.5\% & 5.0\% & 5.6\% & 28.4\%
& 62.8\% & $-3.7$ \\
\midrule

DeepSeek V3.2
& \textbf{51.12} (\textpm 1.5)
& 42.7 (\textpm 1.5)
& 47.3 (\textpm 1.5)
& $\geq$99\% & $\geq$99\% & $\geq$99\% & $\geq$99\%
& N/A & N/A \\

Qwen3.6-Plus
& \textbf{60.20} (\textpm 1.5)
& 48.9 (\textpm 1.5)
& 51.0 (\textpm 1.5)
& 22.7\% & 9.0\% & 14.3\% & 42.9\%
& 56.7\% & $+7.8$ \\

Mistral Large 3
& \textbf{55.33} (\textpm 1.5)
& 32.0 (\textpm 1.4)
& 50.6 (\textpm 1.5)
& 75.0\% & 57.2\% & 76.1\% & 90.0\%
& 44.9\% & $+12.9$ \\
\bottomrule
\end{tabular}
}
\caption{Averaged across models, agentic search improves accuracy by 10.0 percentage points over the search-disabled baseline and by 9.1 points over retrieval-augmented generation (RAG) with one retrieval. Search invocation ranges from 13.5\% to $\geq$99\% and generally increases toward municipal questions. Bold marks the highest accuracy condition within each model. Parentheses report 95\% Wilson confidence-interval half-widths in percentage points. Selective No-Search Accuracy is computed on questions a model chose not to search. $\Delta_{\mathrm{skip}}$ is this value minus the same model's Search-Disabled Accuracy on the full benchmark. $\geq$99\% denotes near-universal search. N/A indicates too few no-search responses.}
\label{tab:search_behavior}
\end{table*}

\subsection{Agentic Search Behavior}
\label{subsec: agentic search}

Accuracy captures whether models answer correctly, but not whether they ground their answers in current authoritative sources or rely on parametric recall. For civic information that changes with policy and regulatory updates, bypassing search risks relying on stale training-time knowledge. Table~\ref{tab:search_behavior} reports search invocation rates and Selective No-Search Accuracy for all ten models.

\paragraph{Default agentic search behavior varies 
substantially across models.}
Search invocation rates range from 13.5\% (Gemini 3 Flash) to 99.8\% (DeepSeek V3.2), an 86-percentage-point spread. The low end includes the proprietary Gemini 3 Flash and the open-source Qwen3.6-Plus, while the high end includes the proprietary Claude variants and the open-source DeepSeek V3.2. These results indicate that default search behavior varies independently of whether a model is open-source or proprietary. For public sector deployment, a model that rarely invokes search relies more on parametric knowledge and may be more exposed to outdated civic information.

\paragraph{Search invocation rate increases substantially with
descending jurisdictional level.}
Federal-to-municipal search invocation gaps (Table~\ref{tab:search_behavior}) range from 23.4 percentage points (Gemini 3 Flash) to 62.0 percentage points (GPT-5.2 Chat). This pattern is consistent with the hypothesis in \S\ref{subsec:results} that local civic information is less represented in general-purpose training corpora, since models with weaker parametric knowledge at lower jurisdictional levels would rely more on search.

\paragraph{Agentic search achieves higher average accuracy than closed-book answering and RAG.}
On average, the ten models score 10.0 percentage points higher with agentic search than with closed-book answering and 9.1 percentage points higher than with single retrieval RAG. Single retrieval RAG improves accuracy by less than 1 percentage point over closed-book answering. Agentic search exceeds single retrieval RAG for every model.

\paragraph{Most models skip search selectively.}
Among the seven models with sufficient numbers of questions answered without search, six have positive $\Delta_{\mathrm{skip}}$ values. Mistral Large 3 has the largest positive value at $+12.9$ percentage points, whereas Gemini 3 Flash is the only model with a negative value at $-3.7$ percentage points.

\subsection{Authoritative Source Citing}

\begin{figure}[t]
  \centering
  \includegraphics[width=\linewidth]{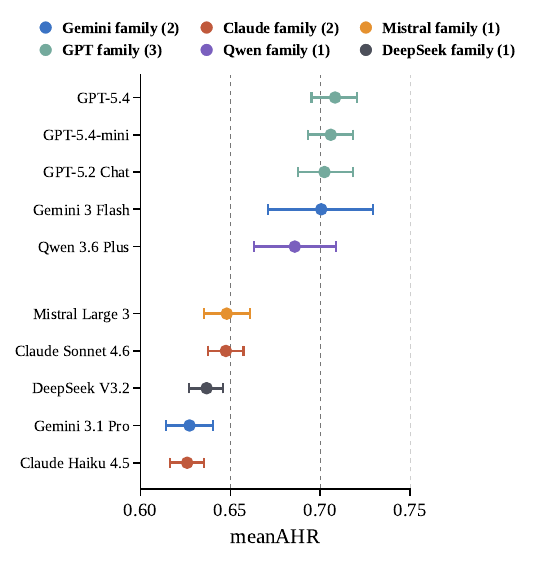}
  \caption{Models partition into two tiers on meanAHR, ordered from highest to lowest. Each point is one model's meanAHR with 95\% bootstrap CI. Points are colored by model family. The GPT family holds the top three positions. Within each tier, CIs overlap.}
  \label{fig:ahr}
  \vspace{-12pt}
\end{figure}

For public-sector deployment, citing authoritative sources ensures the authenticity of the information. In this regard, meanAHR complements prior source-reliability estimation approaches \citep{hwang2025rarag} and citation-grounded evaluation \citep{gao2023alce} by measuring the share of cited URLs from authoritative government domains. Figure~\ref{fig:ahr} demonstrates model-wise meanAHR with 95\% bootstrap confidence intervals.

\paragraph{Models partition into two tiers on meanAHR.}
Models in our evaluation partition into a top tier with meanAHR values of 0.69 to 0.71 and a bottom tier with values of 0.63 to 0.65. A meanAHR of 0.63 indicates that, on the average question, roughly 63\% of the URLs cited by the model are authoritative. Within each tier, confidence intervals overlap, so we treat the two-tier split, which a larger model set might narrow, as the meaningful feature rather than within-tier ordering.

\paragraph{Top-tier performance concentrates within the GPT family.}
Three of the top five models are GPT variants: GPT-5.4, GPT-5.4 mini, 
and GPT-5.2 Chat. These models span different product tiers within 
OpenAI's lineup yet sit within overlapping confidence intervals on 
meanAHR. We hypothesize that GPT-family models share retrieval-grounding 
and citation behaviors that are largely independent of general model 
capability, producing this within-lineup consistency. The meanAHR ordering also diverges from aggregate accuracy. Gemini 3.1 Pro, the strongest model by accuracy (Table~\ref{tab:civi-eval-accuracy}), sits in the lower meanAHR tier, while GPT-5.4-mini ranks in the top meanAHR tier despite placing seventh on accuracy. Rank order on 
meanAHR shows only weak agreement with rank order on LMArena Text Arena
Elo\footnote{LMArena Text-Arena Elo scores accessed from 
\url{https://lmarena.ai/leaderboard} on 2026-05-16.} 
(Spearman $\rho = 0.27$, $p = 0.45$, $N = 10$), with Kendall 
$\tau = 0.24$ ($p = 0.38$) supporting this magnitude. These results 
show insufficient evidence for a correlation between general capability 
rank and meanAHR rank, suggesting that authoritative-source citation 
captures a dimension only weakly aligned with general model capability.

\section{Decomposing Failures}
\label{sec:cafda}

\begin{figure*}[t]
\centering
\includegraphics[width=\textwidth]{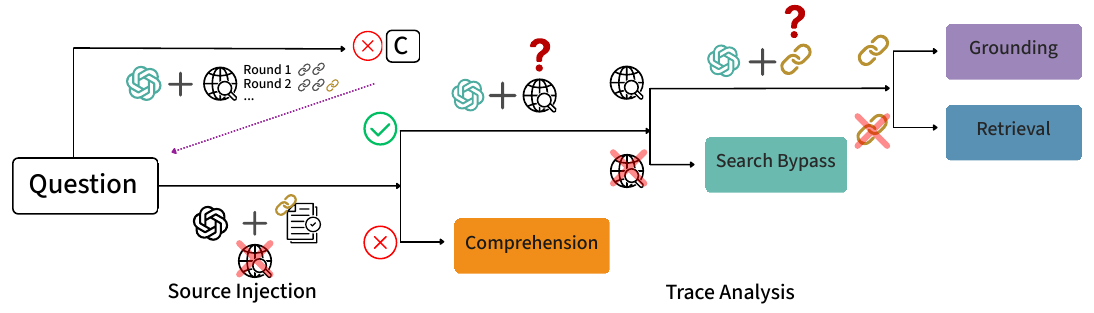}
\caption{The ARISE workflow that categorizes 
incorrect responses from the initial evaluation into 
four failure modes. Specifically, the 
incorrectly answered question is re-evaluated with the gold source page as context with search disabled (\textit{Source Injection}). The failure is categorized as \textit{Comprehension} if the answer remains incorrect in re-evaluation. Otherwise, ARISE investigates whether search is bypassed and whether the gold source page is correctly retrieved in the initial evaluation (\textit{Trace Analysis}).}
\label{fig:arise-diagnostic}
\end{figure*}

The metrics in \S\ref{sec:civi-eval} establish that the evaluated agents
underperform on civic question answering and characterize their search
and sourcing behavior, but they do not distinguish among the ways an
answer can be wrong. We introduce ARISE, CIVI's failure-attribution
procedure, which combines a source-injection ablation following
\citet{cuconasu2024power} with inspection of the original agentic search
trace to deterministically categorize each incorrect response into one
of four failure modes: search bypass, retrieval, grounding, or
comprehension. These modes are mutually exclusive and exhaustive, so
every incorrect response receives exactly one failure label.

\paragraph{Diagnostic methodology.}
We apply ARISE to all observed failures from all ten agents evaluated
in \S\ref{sec:civi-eval}, using the two-step diagnostic described below.

\begin{figure}[!ht]
\centering
\includegraphics[width=\linewidth]{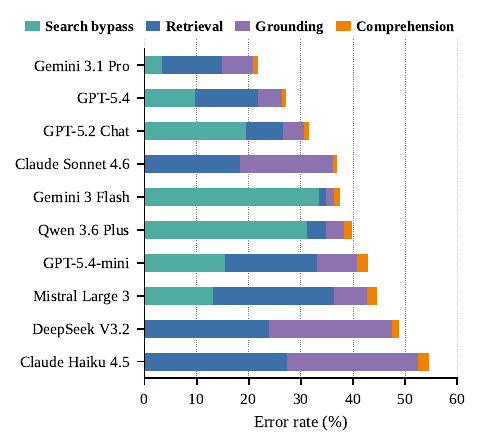}
\caption{Full-population failure diagnostic. Failures are predominantly
retrieval-bound. Search bypass and retrieval failure together account
for an average of 72.1\% of errors across the ten models, suggesting
that most failures occur without the authoritative source in context.
Search bypass accounts for 34.5\% of failures and retrieval failure for
37.6\%. Bar length reflects each model's overall error rate.}
\label{fig:cafda-modes}
\vspace{-10pt}
\end{figure}
 
\paragraph{Step 1: Source injection.}
For each failed response, we rerun the underlying model with search
disabled and supply the gold source page as additional context. This eliminates retrieval, grounding, and search-bypass failure modes by design. If the model still produces an incorrect answer, the failure is attributed to a \textbf{comprehension failure}. An example of this is provided in Appendix~\ref{app:comprehension-example}, where GPT-5.4 cited the verbatim text from the supplied source page but still produced an incorrect answer.

\paragraph{Step 2. Trace analysis.}
For questions where the model now produces the correct answer under source injection, the model demonstrates the ability to comprehend the gold source when supplied directly. We analyze the agentic search trace preserved from the \S\ref{sec:civi-eval} evaluation run. No additional model calls are required, as all trace data was logged during the original evaluation. Three categories are possible. \textbf{Search bypass:} the trace shows no search invocation, indicating the model answered from parametric knowledge alone. \textbf{Retrieval failure:} search was invoked but the gold source URL was not retrieved. \textbf{Grounding failure:} search was invoked and the gold source URL was retrieved, yet the answer was incorrect.

\paragraph{Retrieval-bound failures dominate.}
As shown in Figure~\ref{fig:cafda-modes}, search bypass and retrieval failures together account for an average of 72.1\% of failures across the ten evaluated models, ranging from 48.9\% (DeepSeek V3.2) to 92.9\% (Gemini 3 Flash). Averaged across models, search bypass accounts for 34.5\% of failures and retrieval failure for 37.6\%. These findings indicate that civic information accuracy is most often lost at the retrieval stage, whether through failure to invoke search or failure to surface the authoritative source, rather than at grounding or comprehension. 
 
\paragraph{Improving retrieval cannot eliminate comprehension errors.}
In the full-population diagnostic, between 2.3\% and 4.8\% of each model's failed questions persist even when the gold source page (median \textasciitilde1{,}000 words) is provided in the model's context window (Figure~\ref{fig:cafda-modes}). For example, even when the model is given the gold source page and quotes verbatim text from it, it can still produce an incorrect answer (see Appendix~\ref{app:comprehension-example}). Unlike search bypass, retrieval, and grounding failures, these errors cannot be addressed by better retrieval. This residual may reflect the difficulty of public-sector institutional language, which combines precise terminology with complex conditional structures such as eligibility criteria and regulatory text~\citep{martinez2022poor}. 

\paragraph{Diagnostic implications.}
Because the four modes localize failures at different stages,
they also identify different intervention targets. Search bypass points
to search triggering or civic intent routing, retrieval failure to
query reformulation or authoritative source indexing, grounding failure
to checks for consistency between answers and evidence, and
comprehension failure to stronger reasoning support or human
escalation. These are implications of the diagnostic categories rather
than interventions evaluated in this study.

\section{Related Work}

\paragraph{Failure Decomposition.}
Diagnosing failures in retrieval-augmented and 
agentic LLM systems has been approached through 
trajectory-step taxonomies, claim-level metrics, and 
pipeline-stage 
decompositions~\citep{jiao2026doctorrag, 
sivakumar2026ragx, ru2024ragchecker, zhang2025deft, 
leung2026classifying, shao2025seekbench}.
\citet{mallen2023whennot} establish the 
parametric-versus-non-parametric memory distinction 
that motivates source-injection ablation, while 
\citet{wu2025searchwisely} taxonomize suboptimal 
agentic search behavior into over-search and 
under-search categories. To our knowledge, no prior work applies source-injection ablation together with agentic search trace analysis to civic information. ARISE combines them to decompose failures into search bypass, retrieval, grounding, and comprehension.

\paragraph{LLM Evaluation for Civic and Public-Sector Information.}
A growing line of benchmarks evaluates LLM accuracy on government information at single-country, single-jurisdictional scopes \citep{majithia2026citizenquery, harris2025pubhealthbench, gao2025localbench, liu2025evaluation, yang2025polis, wang2025govrelbench}. Some benchmarks also evaluate municipal open-data and regulatory knowledge \citep{siebenmann2026ogd4all, guha2023legalbench}. CIVI's benchmark instantiation is the first to jointly span cross-national and inter-jurisdictional civic QA under an
international functional classification standard, covering three
federal democracies, three jurisdictional levels, and four COFOG categories.

\section{Conclusion}

We present CIVI, the first framework for diagnosing search agent
failures in civic information, instantiated across three federal
democracies, three jurisdictional levels, and functional categories
from an internationally adopted United Nations standard. CIVI performs
failure attribution through ARISE, which combines source-injection
ablation with agentic search trace analysis to distinguish search
bypass, retrieval, grounding, and comprehension failures. None of the
ten evaluated agents matches the attentive human baseline, and ARISE
attributes 72.1\% of failures to retrieval-bound causes, comprising
34.5\% search bypass and 37.6\% retrieval failure. By localizing
failures to different stages, CIVI indicates whether interventions
should target search invocation, authoritative source retrieval,
answer-evidence consistency, or comprehension support and human
escalation. More broadly, civic deployers should consider not only
accuracy but also how quickly information changes, the jurisdictional
level served, and whether answers must be traceable to authoritative
government sources.

\section*{Limitations}

\textbf{Geographic and linguistic scope.} The present evaluation
focuses on English-language civic information across three countries
(Canada, the United States, and Australia). Although Canada is
officially bilingual, our source pages and questions are in English
only. The COFOG based construction protocol is country agnostic, but expansion requires experts familiar with each country's languages and administrative system. Extending CIVI to multilingual civic QA and to governance
systems with different legal or administrative traditions is a natural
direction for future work.

\textbf{Uniform search substrate.} All models access agentic search
through a single provider (Exa), a design choice that prioritizes
controlled comparison across model families. Future evaluations could
complement this by testing models under their native search
infrastructure to assess ecological validity in specific deployment
settings.

\section*{Ethical Considerations}

\paragraph{Human participants.}
The human baseline used as a human comparator in our evaluation
involved five volunteer participants. Participants were recruited  from the authors' personal networks and did not
receive compensation. Participation was entirely voluntary, and participants
were free to withdraw at any time. Before starting, each participant
was informed about the purpose of the study, the nature of the
question-answering task, and its expected duration, and gave informed
consent. We collected no personally identifiable or sensitive
information.  The task consisted of
answering publicly available civic questions and posed
minimal risk to participants.

\bibliography{references}

\FloatBarrier
\clearpage

\appendix

\section{Dataset Format}
\label{app:format}

We release the CIVI dataset under the Creative Commons Attribution 4.0 International (CC BY 4.0) license and the accompanying code under the MIT License. The underlying source pages are drawn from government materials available under open-government provisions, including the Open Government Licence (Canada and Australia), U.S. federal public-domain works, or comparable open-access provisions at the state and municipal level.

\begin{table*}[t]
\centering
\small
\caption{QA pair counts per stratification cell. CIVI
comprises 4,097 QA pairs across 36 cells, formed by 4 COFOG categories
$\times$ 3 countries $\times$ 3 jurisdictional levels.}
\label{tab:cell-counts}
\begin{tabular}{lrrrrr}
\toprule
Jurisdiction
 & \shortstack{03 Public\\Order \& Safety}
 & \shortstack{04 Economic\\Affairs}
 & \shortstack{07\\Health}
 & \shortstack{10 Social\\Protection}
 & \textbf{Total} \\
\midrule
Australia, Federal       &  70 & 114 & 113 & 121 & \textbf{418} \\
Australia, State         & 127 & 112 & 100 &  99 & \textbf{438} \\
Australia, Municipal     & 142 & 101 & 107 & 122 & \textbf{472} \\
\addlinespace
Canada, Federal          &  82 & 137 & 100 & 110 & \textbf{429} \\
Canada, Provincial       & 104 &  98 &  88 & 115 & \textbf{405} \\
Canada, Municipal        & 140 & 143 &  75 & 134 & \textbf{492} \\
\addlinespace
United States, Federal   &  96 & 114 & 134 & 119 & \textbf{463} \\
United States, State     & 133 & 146 & 106 & 106 & \textbf{491} \\
United States, Municipal & 115 & 145 &  84 & 145 & \textbf{489} \\
\midrule
\textbf{Column total} & \textbf{1,009} & \textbf{1,110} & \textbf{907}
 & \textbf{1,071} & \textbf{4,097} \\
\bottomrule
\end{tabular}
\end{table*}

\begin{figure*}[t]
\centering
\footnotesize
\begin{verbatim}
{
  "country": "Australia",
  "layer": "Federal",
  "institutions": [
    {
      "institution": "Australian Government",
      "short_name": "australian_government",
      "pages": [
        {
          "url": "https://www.acic.gov.au/i-need-check-myself",
          "cell": "03",
          "cell_name": "Public Order and Safety",
          "label": "Apply for a police check"
        },
        {
          "url": "https://www.acic.gov.au/how-service-works",
          "cell": "03",
          "cell_name": "Public Order and Safety",
          "label": "How the National Police Checking Service works"
        }
      ]
    }
  ]
}
\end{verbatim}
\caption{Excerpt of the curated source index. Source pages are organized
by country, jurisdictional layer, and institution, with each page
assigned to one COFOG category. The full index covers 576 pages across
36 strata (4 COFOG categories $\times$ 3 countries $\times$
3 jurisdictional levels).}
\label{fig:source-index}
\end{figure*}

\subsection{Source Page Curation}
\label{app:source-curation}

Source pages are organized by country, jurisdictional layer, and
institution, with each page assigned to one COFOG category. The full
index covers 576 pages across 36 strata. An excerpt of the index,
showing one institution, is given in Figure~\ref{fig:source-index}.

\subsection{Example QA Instance}
\label{app:example-qa}

Each evaluation instance is grounded in a single curated source page.
The instance below is drawn from the police check page listed in
Figure~\ref{fig:source-index}.

\smallskip
\noindent\textbf{Question ID:} \texttt{australian\_government\_q001} \\
\textbf{Institution:} Australian Government \\
\textbf{Jurisdiction:} Federal \\
\textbf{COFOG category:} Public Order and Safety \\
\textbf{Source page:} \emph{Apply for a police check}, \\
\url{https://www.acic.gov.au/i-need-check-myself}

\smallskip
\noindent\textbf{Question.} How can an individual obtain a Nationally
Coordinated Criminal History Check? (Select all that apply)

\begin{itemize}\setlength\itemsep{1pt}
  \item[A.] Apply through a state or territory Screening Unit
  \item[B.] Apply online through the National Police Checking Service
            Support System (NSS) directly
  \item[C.] Submit an application to an ACIC accredited body
  \item[D.] Submit an application directly to the ACIC
  \item[E.] Submit an application through the Australian Federal Police
            website only
  \item[F.] Submit an application through the Department of Home Affairs
  \item[G.] Apply through any government agency that handles identity
            verification
  \item[H.] Apply through an Australian police agency
\end{itemize}

\noindent\textbf{Gold answer:} \{C, H\}

\smallskip
\noindent\textbf{Supporting evidence.} Verbatim spans from the source page:
\begin{quote}
``The ACIC does not accept applications or submit checks on behalf of
individuals.''

\smallskip
``To get a check, you can either:\\
submit an application to an ACIC accredited body\\
apply through an Australian police agency.''
\end{quote}

\FloatBarrier

\section{Comprehension Error Example}
\label{app:comprehension-example}

The following illustrates a comprehension failure 
by GPT-5.4 under source injection.

\medskip
\noindent\textbf{Question:} In Australia, under what 
circumstances can a child attending secondary school 
still be eligible for the Child Care Subsidy? 
(Select all that apply)

{\small
\noindent A.~The child is aged under 15 years or has 
a medical condition \quad B.~The child is aged 13 
years of age or under \quad C.~The child is aged 
14--18 years and has a disability \quad D.~The child 
is in foster care \quad E.~The child attends an 
approved after-school program \quad F.~The child is 
aged 14--18 years and the family income is below a 
threshold \quad G.~The child is aged under 16 years 
regardless of disability status \quad H.~The child 
is aged 13--18 years and needs supervision
}

\smallskip
\noindent\textbf{Correct answer:} B, C

\smallskip
\noindent\textbf{Gold source URL:} 
\url{https://www.servicesaustralia.gov.au/who-can-get-child-care-subsidy}

\smallskip
\noindent\textbf{Relevant source text:}
\begin{quote}
``If your child attends secondary school, you may 
still be eligible if they need supervision. They 
must be either: 13 or under; 14 to 18 with 
disability.''
\end{quote}

\noindent\textbf{GPT-5.4 response (verbatim):}
\begin{quote}
The correct answers are C and H.

In Australia, a child attending secondary school 
can still be eligible for Child Care Subsidy if 
they need supervision and are either:
\begin{itemize}
\item 13 or under
\item 14 to 18 with disability
\end{itemize}
So:
\begin{itemize}
\item C.~The child is aged 14--18 years and has a 
disability \checkmark
\item H.~The child is aged 13--18 years and needs 
supervision \checkmark, but more precisely: 13 or 
under, or 14--18 with disability, and they need 
supervision.
\end{itemize}
\end{quote}

\noindent\textbf{Analysis.} The source states that 
supervision is a prerequisite for all secondary-school 
children, with two age-based conditions: 13 or under 
(B), or 14--18 with a disability (C). GPT-5.4 
correctly identified the disability condition (C) and 
reproduced the source text accurately in its 
reasoning, but conflated the supervision prerequisite 
with the age range to select H (``13--18 years and 
needs supervision'') while missing option B (``13 
years of age or under''), which maps directly to the 
source's ``13 or under'' criterion. The model's own 
explanation acknowledges the correct conditions but 
its final selection does not reflect them. The full source page is approximately 800 words. The relevant excerpt is shown above.

\section{Additional Robustness Analyses}
\label{app:robustness}

\subsection{Matched Comparison on Human Questions}
\label{app:matched-human}

Table~\ref{tab:matched-human} compares each model with the human baseline on the same 150 questions. Gemini 3.1 Pro, the strongest model on this subset, is 15.3 percentage points below the human baseline. Across models, the mean absolute difference between accuracy on the matched subset and the full benchmark is 1.8 percentage points.

\begin{table}[t]
\centering
\footnotesize
\caption{Accuracy on the 150 questions answered by the human participants and on the full 4{,}097 question benchmark. N/A indicates that the human participants were evaluated only on the matched subset.}
\label{tab:matched-human}
\begin{tabular}{lcc}
\toprule
\textbf{Model} & \textbf{Matched} & \textbf{Full} \\
\midrule
Human baseline     & 92.67 & N/A \\
Gemini 3.1 Pro     & 77.33 & 78.28 \\
GPT-5.4            & 71.33 & 72.88 \\
GPT-5.2 Chat       & 70.00 & 68.46 \\
Claude Sonnet 4.6  & 61.33 & 63.05 \\
Gemini 3 Flash     & 62.00 & 62.44 \\
Qwen3.6-Plus       & 56.67 & 60.20 \\
GPT-5.4-mini       & 60.67 & 57.18 \\
Mistral Large 3    & 52.00 & 55.33 \\
DeepSeek V3.2      & 49.33 & 51.12 \\
Claude Haiku 4.5   & 45.33 & 45.45 \\
\bottomrule
\end{tabular}
\end{table}

\subsection{Country Results}
\label{app:country-results}

Table~\ref{tab:country-results} reports the country results averaged equally across the ten models.

\begin{table}[t]
\centering
\footnotesize
\caption{Accuracy averaged equally across the ten models by country.}
\label{tab:country-results}
\begin{tabular}{lrr}
\toprule
\textbf{Country} & \textbf{Questions} & \textbf{Accuracy} \\
\midrule
United States & 1{,}443 & 63.4\% \\
Australia     & 1{,}328 & 62.0\% \\
Canada        & 1{,}326 & 59.1\% \\
\bottomrule
\end{tabular}
\end{table}

\subsection{Stability Across Three Runs}
\label{app:stability}

Table~\ref{tab:stability} reports three runs on the same stratified subset of 410 questions. The mean within model range between the maximum and minimum is 2.2 percentage points and the largest is 3.4 percentage points. Only Gemini 3 Flash and Qwen3.6-Plus exchange order, and their full benchmark confidence intervals in Table~\ref{tab:civi-eval-accuracy} overlap.

\begin{table*}[t]
\centering
\footnotesize
\caption{Accuracy across three runs on the same stratified subset of 410 questions. Range is the maximum minus the minimum in percentage points.}
\label{tab:stability}
\begin{tabular}{lrrrr}
\toprule
\textbf{Model} & \textbf{Run 1} & \textbf{Run 2} & \textbf{Run 3} & \textbf{Range} \\
\midrule
Gemini 3.1 Pro    & 78.54 & 78.05 & 78.54 & 0.49 \\
GPT-5.4           & 73.66 & 72.44 & 74.39 & 1.95 \\
GPT-5.2 Chat      & 69.02 & 68.29 & 68.29 & 0.73 \\
Claude Sonnet 4.6 & 63.90 & 64.88 & 62.20 & 2.68 \\
Gemini 3 Flash    & 61.71 & 63.41 & 60.49 & 2.92 \\
Qwen3.6-Plus      & 59.27 & 61.22 & 61.46 & 2.19 \\
GPT-5.4-mini      & 58.54 & 59.27 & 57.56 & 1.71 \\
Mistral Large 3   & 56.59 & 53.66 & 57.07 & 3.41 \\
DeepSeek V3.2     & 49.27 & 52.00 & 50.49 & 2.73 \\
Claude Haiku 4.5  & 44.63 & 43.17 & 46.10 & 2.93 \\
\bottomrule
\end{tabular}
\end{table*}

\FloatBarrier

\section{Implementation Details}
\label{app:implementation}

\subsection{QA Pair Generation}
\label{app:gen-prompt}

All QA pairs were generated with Claude Opus 4.6
(\texttt{anthropic/claude-opus-4.6}) through OpenRouter at
temperature 0. Each source page was supplied in full in one call,
without chunking or retrieval. The system prompt is shown in
Figure~\ref{fig:gen-prompt}, and the user message contains the
institution name, country, and jurisdictional level, followed by the
page text. Generation follows a fixed JSON schema. The model generates
3--10 eight-option questions per page according to content density and
returns verbatim quotations supporting each correct answer.

Generated questions pass two automatic checks before human validation.
A structural check requires eight uniquely labeled options and at least
one correct answer. Malformed questions are dropped, and generation is
attempted at most three times for any page whose output fails to parse
or validate. A quote check verifies every supporting quotation against
the source page by substring matching after whitespace normalization.
Pairs with missing or unmatched quotations are discarded, while
retained quotations are stored verbatim. After both checks, 4{,}354 QA
pairs remained for human validation (\S\ref{subsec:validation}).

\begin{figure*}[t]
\centering
\footnotesize
\fbox{%
\begin{minipage}{0.97\textwidth}
You are a question writer for a factual knowledge benchmark. You will be
given the full text of a web page. Generate between \{min\_questions\}
and \{max\_questions\} questions. Decide how many based on how much
content is on the page --- more content means more questions, less
content means fewer.

\smallskip
Generate a balanced set of source-grounded questions that evenly covers
the page's major decision-relevant content. The questions should span
distinct aspects of the service rather than clustering around the same
section, detail, or information type. When present on the page,
distribute questions across different decision-relevant information
types, such as eligibility, required documents, deadlines, fees,
application or service steps, exceptions, locations/contact channels,
processing times, restrictions, and emergency or high-risk instructions.
Treat these as coverage prompts rather than mandatory categories. Avoid
redundant or near-duplicate questions.

\smallskip
All questions must be multiple-choice (MC).

\smallskip
\textbf{Multiple-choice (MC) rules}
\begin{itemize}\setlength\itemsep{2pt}
  \item Generate \{num\_options\} options (labels A--\{max\_label\}).
  \item End the question with ``(Select all that apply)''.
  \item There may be one or more correct answers.
  \item Construct each distractor in one of two ways:
    \begin{itemize}\setlength\itemsep{1pt}
      \item[(a)] Take real content from the page and apply a controlled
        perturbation. Examples include but are not limited to: changing
        a number, swapping a threshold, flipping a condition, altering a
        deadline, substituting a nearby-but-incorrect name, or reordering
        a process step.
      \item[(b)] Construct a plausible misconception that a reader might
        believe based on common assumptions or related civic knowledge,
        but that is not supported by the source page.
    \end{itemize}
  \item Each distractor must test a distinct misconception. Do not
    generate multiple distractors that make the same claim with different
    wording. If two distractor candidates would be eliminated by the same
    reasoning, combine them or replace one with a different perturbation
    type.
  \item Each question should ask about information that has consequences
    for residents interacting with services. Based on the
    nature of the page, choose the appropriate question type:
    \begin{itemize}\setlength\itemsep{1pt}
      \item[(a)] Action-relevant questions: who qualifies (eligibility),
        what they pay (fees), when they must act (deadlines), what they
        need to provide (documentation), how they should proceed
        (process), where they need to go (channels), what they receive
        (entitlements). Use these for pages about applying for services,
        eligibility rules, or process descriptions.
      \item[(b)] Transparency questions: how the service operates, what's
        being delivered, how the system works in ways residents would
        reasonably want to verify or understand. Use these for pages
        describing how a service works, or how
        performance is measured.
    \end{itemize}
    A single page may support both types --- generate the mix that the
    page content naturally supports. Avoid questions whose answer is
    purely descriptive trivia with no practical or accountability
    significance.
  \item DO NOT ask trivial definition or abbreviation questions.
    (Definitional questions about what a bylaw covers, who or what
    qualifies as regulated, or scope of eligibility are not trivial.)
  \item DO NOT ask questions whose answer is obvious from common
    knowledge.
  \item Prefer questions that combine or compare multiple pieces of
    information, testing whether the reader understands how they relate.
  \item Do NOT start the question with ``Which of the following
    statements'' or ``Which of the following''.
  \item Write questions the way a real person would ask them --- direct
    and specific. Vary the phrasing naturally. Do not repeat the same
    question structure across multiple questions.
  \item Do NOT use generic phrasing like ``accurately describe'' or
    ``correctly describe''. Be specific about what you are asking.
\end{itemize}

\smallskip
\textbf{Quotes.} Include verbatim quotes from the source text that
support the correct answer(s). Copy quotes exactly as they appear in the
source text. Preserve line breaks. Do not convert bullet points or
numbered lists into comma-separated text. Do not truncate with ellipsis.
Do not paraphrase.

\smallskip
\rule{\linewidth}{0.4pt}
\smallskip

\textbf{User prompt.} \{institution\_header\}Source text:
\{page\_text\}

\smallskip
Return JSON matching the schema exactly.
\end{minipage}}
\caption{Prompt used to generate multiple-choice QA pairs from each
curated source page. Placeholders in braces are filled at generation
time.}
\label{fig:gen-prompt}
\end{figure*}

\subsection{Evaluation}
\label{app:eval}

All ten evaluated models were accessed through the OpenRouter API and queried at temperature 0. Agentic search is provided to each model as
a tool that issues queries to the Exa search API, which the model may
invoke while answering. Following the retrieval budget in
\S\ref{subsec:eval-setup}, each model may issue up to three rounds of
search queries and retrieve at most ten URLs in total. For scoring, the selected
option letters are extracted from each model's final response by a
separate model, Gemini 2.5 Flash, and the extracted set is compared
against the gold answer set. A response is counted correct only when
the extracted set matches the gold set exactly, with no missing and no
additional letters. To verify this extraction step, the extracted
answers were manually checked against the corresponding model
responses for a sample of 100 questions, and all 100 were correct. For each
question, the model's search queries, the retrieved URLs, the cited
URLs, and the final response are logged.

\subsection{Search Baseline Conditions}
\label{app:baseline-details}

The search disabled and single retrieval RAG baselines use the same ten model versions, all 4{,}097 questions, the same eight option select all response format, temperature 0, and the same exact match scoring procedure as the agentic condition. In the search disabled condition, search is unavailable and the model receives no retrieved context. In single retrieval RAG, the verbatim question is issued once to the same Exa search substrate, and the top three results are inserted into the model context. This condition includes no query reformulation, no model decision about whether to search, and no additional retrieval round. Table~\ref{tab:search_behavior} reports 95\% Wilson score confidence intervals for both baselines.

\section{Source Page Curation Criteria}
\label{app:curation}
 
Candidate source pages were drawn exclusively from official government domains (e.g., \texttt{.gc.ca}, \texttt{.gov}, \texttt{.gov.au}, and accredited subdomains for state, provincial, and municipal authorities). Three domain experts with public-sector experience across research, practitioner, and advisory roles curated the source pages, retaining only pages that satisfied four criteria:
 
\textbf{Curation procedure.} The 576 source pages were curated by three domain experts in public-sector information, one cell at a time. For each of the 36 cells, an expert first determined which government body is responsible for the cell's function at its jurisdictional level, a mapping decided during curation rather than fixed in advance. The expert then worked through the body's official website, navigating to the service categories matching the function and reviewing many pages within them as candidates. For example, an expert curating a local cell for Canada reviewed the snow-clearing service pages on the City of Toronto website. For each cell this produced a candidate pool of at least four times the 16-page target.
\textbf{Selection criteria.} From each candidate pool, pages were retained according to four criteria.
\textbf{Authoritative source.} The page is hosted on an official government domain or on the domain of a government-funded public-sector body.
\textbf{Substantive content.} The page presents substantive content rather than functioning as a landing page or navigation hub. Retained pages have a median length of roughly 1{,}000 words, and shorter pages were kept only when dense with information relevant to a resident's decision.
\textbf{Resident relevance.} The content reflects information residents would need for a real task or decision.
\textbf{No AI-use restriction.} To the best of the experts' determination, the page carries no terms restricting its use for AI training or evaluation. A small number of pages were excluded on this basis.
\textbf{Page count and finalization.} The target of 16 pages per cell was fixed in advance. Most cells admitted well over 16 qualifying pages, so curation primarily involved trimming a larger qualified set to the 16 most substantive and representative pages. One expert led candidate gathering, and all three experts reviewed and finalized the selection for every cell together.

\end{document}